\documentclass[conference,a4paper]{IEEEtran}
\IEEEoverridecommandlockouts

\usepackage{hyperref}
\hypersetup{hidelinks}
\usepackage[cmex10]{amsmath}%American Math Society(AMS) math formatting
\usepackage{amssymb,amsfonts}%AMS extra symbols and fonts
\usepackage{dblfloatfix}%fix double column figure ordering and placement

\usepackage[ruled,vlined]{algorithm2e}
\usepackage{circledsteps}
\usepackage[subpreambles=true]{standalone}
\usepackage{graphicx}
\usepackage{multirow}
\usepackage{makecell}
\usepackage{enumitem}
\graphicspath{{Figures/PDF/}{Figures/PNG/}}
\usepackage{acronym}
\usepackage{placeins}
\usepackage{bbold}
\usepackage{dsfont}
\usepackage{cleveref}
\Crefname{figure}{Fig.}{Figs.}
\usepackage{caption}
\usepackage{subcaption}

\usepackage{booktabs}
\usepackage{siunitx}
\usepackage[natbib=true,style=ieee,sorting=none,maxnames=4]{biblatex}
\usepackage{texnames}
\usepackage{bm,bbm}
\usepackage{orcidlink}

\usepackage{float}
\usepackage{pifont}
\usepackage{csquotes}
\usepackage{flushend}

\newcommand{\cmark}{\ding{51}}
\newcommand{\xmark}{\ding{55}}

\newcommand{\red}[1]{\textcolor{red}{#1}}
\newcommand{\green}[1]{\textcolor{teal}{#1}}
\newcommand{\bskip}{-14pt}
\newcommand{\ours}[1]{#1$^*$}
\newcommand{\oursnumber}[1]{#1}
\newcommand{\new}[1]{#1}

\DeclareMathOperator{\att}{att}
\DeclareMathOperator{\FC}{FC}
\DeclareMathOperator{\TE}{TransformerEncoder}
\DeclareMathOperator{\bleu}{BLEU}
\DeclareMathOperator{\uni}{u}

\begin{document}

\acrodef{adamw}[AdamW]{Adam with decoupled weight decay}
\acrodef{ap}[AP]{average precision}
\acrodef{bert}[BERT]{bidirectional encoder representations from transformers}
\acrodef{bleu}[BLEU]{bilingual evaluation understudy}
\acrodef{cbir}[CBIR]{content-based image-retrieval}
\acrodef{clip}[CLIP]{contrastive language-image pretraining}
\acrodef{cnn}[CNN]{convolutional neural network}
\acrodef{cv}[CV]{computer vision}
\acrodef{flops}[FLOPS]{floating point operations per second}
\acrodef{fm}[FM]{foundation model}
\acrodef{llm}[LLM]{large language model}
\acrodef{lfa}[LFA]{learning-based feature aggregation}
\acrodef{lstm}[LSTM]{long short-term memory}
\acrodef{map}[mAP]{mean average precision}
\acrodef{mlm}[MLM]{masked language modelling}
\acrodef{mlp}[MLP]{multi-layer perceptron}
\acrodef{mtld}[MTLD]{measure of textual lexical diversity}
\acrodef{mv-vlm}[MV-VLM]{multi-viewpoint \ac{vlm}}
\acrodef{ndcg}[NDCG]{normalized discounted cumulative gain}
\acrodef{ner}[NER]{next entity recognition}
\acrodef{nlp}[NLP]{natural language processing}
\acrodef{ntxent}[NT-Xent]{normalized temperature-scaled cross entropy}
\acrodef{rs}[RS]{remote sensing}
\acrodef{rsicap}[RSICap]{remote sensing image captioning dataset}
\acrodef{sift}[SIFT]{scale-invariant feature transform}
\acrodef{ttr}[TTR]{type-token ratio}
\acrodef{vlm}[VLM]{vision-language model}
\acrodef{lgwf}[LG-WF]{Localized Weight Generation}
\acrodef{wfa}[WFA]{weighted feature aggregation}

\title{
     \scshape Redundancy-Aware Pretraining of Vision-Language Foundation Models in Remote Sensing
}

\author{
    \IEEEauthorblockN{Mathis Jürgen Adler\orcidlink{0009-0007-1865-1492}$^{*1,2}$, %
    Leonard Hackel\orcidlink{0000-0002-5831-1237}$^{*1,2}$, %
    Gencer Sumbul\orcidlink{0000-0003-3690-3052}$^{3}$,
    Begüm Demir\orcidlink{0000-0003-2175-7072}$^{1,2}$}\\
    
    \IEEEauthorblockA{
    \hfill
    $^1$\,TU Berlin, Germany \hfill $^2$\,BIFOLD, Germany \hfill
    $^3$\,EPFL, Switzerland
    \hfill
    }
}

\maketitle
\def\thefootnote{*}\footnotetext{These authors contributed equally to this work}\def\thefootnote{\arabic{footnote}}

\begin{abstract}
The development of foundation models through pretraining of \acp{vlm} has recently attracted great attention in \ac{rs}. 
\ac{vlm} pretraining aims to learn image and language alignments from a large number of image-text pairs. 
Each pretraining image is often associated with multiple captions with redundant information due to repeated or semantically similar phrases that result in increased pretraining and inference time of \acp{vlm}.
To overcome this, we introduce a \ac{wfa} strategy for \ac{vlm} pretraining in \ac{rs}. Our strategy aims to extract and exploit complementary information from multiple captions per image, while reducing redundancies through feature aggregation with importance weighting. 
To calculate adaptive importance weights for different captions of each image, we propose two different techniques: i) non-parametric uniqueness; and ii) learning-based attention. 
In the first technique, importance weights are calculated based on the \ac{bleu}-scores of the captions to emphasize unique sentences while removing the influence of repetitive sentences. 
In the second technique, importance weights are learned through an attention mechanism instead of relying on hand-crafted features. 
The effectiveness of the proposed \ac{wfa} strategy with the two techniques is analyzed in terms of downstream performance on text-to-image retrieval in \ac{rs}. 
Experimental results show that the proposed strategy enables efficient and effective pretraining of \acp{vlm} in \ac{rs}. Based on the experimental analysis, we derive guidelines for the proper selection of techniques, considering downstream task requirements and resource constraints. 
\new{The code of this work is publicly available at \url{https://git.tu-berlin.de/rsim/redundacy-aware-rs-vlm}.}
\end{abstract}


\acresetall  % reset all acronyms
\section{Introduction}
Inspired by the success of \ac{llm}-based \acp{fm} in \ac{nlp} such as LLama \cite{touvron2023llama} and GPT-3 \cite{brown2020language}, \acp{vlm} in \ac{rs} have recently exhibited substantial advancements \cite{rs16091477, liu2024remoteclip, hu2023rsgptremotesensingvision, zhang2024earthgpt}.
The \acp{vlm} learn image-language alignments from a large number of image-text (i.e., image-caption) pairs and are then fine-tuned with a small amount of labeled data. 
Existing \acp{vlm} in \ac{rs} demonstrate remarkable capability in several image-language tasks, such as image-text retrieval \cite{liu2024remoteclip, rs15184637}, visual question answering \cite{hu2023rsgptremotesensingvision}, zero-shot classification \cite{zhang2024earthgpt} and image captioning \cite{hu2023rsgptremotesensingvision}. 

For example, in \cite{liu2024remoteclip}, continual pretraining of a \ac{clip} model for \ac{rs} alignment is presented, showing that the \ac{rs}-specific alignment improves performance over the domain-agnostic \ac{clip} on various \ac{rs}-specific segmentation, detection and retrieval tasks.

In \cite{hu2023rsgptremotesensingvision}, a small-scale, albeit high quality \ac{rs} image captioning dataset is constructed to fine-tune a \ac{vlm} for \ac{rs}. 
In this work, it is shown that pretraining on small-scale high-quality data can improve the performance of \ac{rs}-\acp{vlm} in downstream tasks (such as captioning and visual question answering) compared to pretraining on large but automatically annotated datasets with lower variety and quality.
For a comprehensive overview of \acp{vlm} in \ac{rs}, we refer the reader to \cite{VlmsInRsOverview}.

\begin{figure}
    \centering
    \begin{minipage}{0.14\textwidth}
        \includegraphics[width=\textwidth]{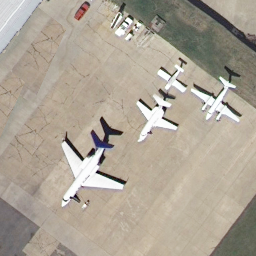}
    \end{minipage}
        \hspace*{-8mm}
    \hfill
    \begin{minipage}{0.35\textwidth}
    \footnotesize
    \begin{enumerate}[itemsep=0.2cm]
        \item \red{Four airplanes} are \green{parked} \red{at the airport}.
        \item There are \red{some planes} and \red{cars in the airport}.
        \item \red{Four different} \green{kinds} of \red{airplanes} are \red{in the airport}.
        \item \red{Four different} \green{sizes} of \red{airplanes} are \red{in the airport}.
        \item Here are \red{some airplanes} and \red{cars in the airport}.
    \end{enumerate}
    \end{minipage}
    \captionsetup{belowskip=\bskip}
    \caption{An example image from the UCM-captions dataset \cite{qu2016deep} with highly redundant (shown in \red{red}) and unique (shown in \green{teal}) information within its five captions.}
    \label{fig:ucm_sample}
\end{figure}

To pretrain \acp{vlm} in \ac{rs}, various datasets have recently been developed and made publicly available \cite{liu2024remoteclip, zhang2024earthgpt, hu2023rsgptremotesensingvision}.
Most \new{large-scale \ac{vlm} pretraining} datasets include existing image captioning datasets such as UCM-captions \cite{qu2016deep} and NWPU-Captions \cite{cheng2022nwpu}.
\Cref{fig:ucm_sample} shows an example image from the UCM-captions dataset that contains five captions per image.
The five sentences share almost the same information, with one sentence (sentence 5) containing no unique information at all. 
Although existing \acp{vlm} in RS have demonstrated remarkable potential in aligning image and text pairs, they can not efficiently handle redundant information. 
A widely used strategy to exploit such data in \ac{rs} \ac{vlm} pretraining is to replicate each image as many times as the number of its captions (denoted as Replication strategy in this paper) \cite{liu2024remoteclip, rs16091477, zhang2024earthgpt}. 
We would like to note that, when a model processes all the image caption pairs individually, redundant information has to be processed multiple times. 
Thus, this strategy results in an increase in the pretraining and inference time of \acp{vlm} (without having any benefit in the performance of the downstream tasks).

To address this issue, in this paper we present a \ac{wfa} strategy for \ac{vlm} pretraining in \ac{rs}.
Our strategy exploits complementary information from all captions of an image, while reducing redundancies through feature aggregation with importance weighting. 
For importance estimation, we introduce two techniques: i) non-parametric uniqueness; and ii) learning-based attention.

We compare both techniques of our \ac{wfa} strategy with several baseline strategies, and derive some guidelines for their use.

\section{Methodology}\label{sec:methodology}

Let  $\mathcal{X} = \{(I_i, \mathcal{C}_i)\}_{i=1}^N$ be a pretraining set including $N$ pairs, where $I_i$ is the $i$th image and $\mathcal{C}_i$ is the set of captions associated to $I_i$. 
Each $\mathcal{C}_i$ contains $M_i$ individual captions, i.e., $\mathcal{C}_i = \left\{c_{i, 1}, c_{i, 2}, \dots, c_{i, M_i}\right\}$, where $c_{i,j}$ is the $j$th caption that is associated with image $I_i$.
For the sake of simplicity, $\mathcal{C}_i$, $I_i$, $M_i$ and $c_{i,j}$ are denoted as $\mathcal{C}$, $I$, $M$ and $c_j$, respectively, in the rest of the paper.
\ac{vlm} pretraining is generally achieved by \ac{clip} \cite{radford2021learning} that employs a text encoder $\phi_{\mathcal{C}}$, an image encoder $\phi_I$ and a projection head $\phi_p$. 
For a given pretraining image $I$, the image encoder extracts image features $H_I = \phi_I(I)$, while the \ac{llm}-based text encoder extracts text features $H_{c_j} = \phi_{\mathcal{C}}(c_j)$ from each caption $c_j$ individually and a projection head employs an additional feature alignment step. 
The \ac{ntxent} loss function $\mathcal{L}_{\text{NTX}}$~\cite{sohn2016improved} is utilized to model image-text alignment.

We assume that the captions in $\mathcal{C}$ can contain not only partially complementary but also redundant information. 
This increases the complexity of \ac{vlm} pretraining if $I$ is replicated $M$ times as in the widely used Replication strategy. 
As an alternative to the Replication strategy, the mean of caption features can be associated with the image feature, where multiple captions contribute equally to the overall representation. 
However, as shown in \Cref{fig:ucm_sample}, certain captions may be partially or entirely redundant, and thus provide limited descriptive value. To overcome this, we propose a \ac{wfa} strategy for \ac{vlm} pretraining that utilizes feature aggregation based on the importance of all captions in $\mathcal{C}$ weighted with respect to their informational content. 
This is achieved by weighted averaging of text features as follows:
\begin{equation}\label{eq:weighted_avg}
    H_{\mathcal{C}} = \sum_{j=1}^M \alpha_j~\cdot H_{c_j},
\end{equation}
where $H_\mathcal{C}$ is the aggregated text feature and $\alpha_j$ is the weighting factor for caption feature $H_{c_j}$. 

\begin{figure}
    \centering
    \includegraphics[width=\linewidth]{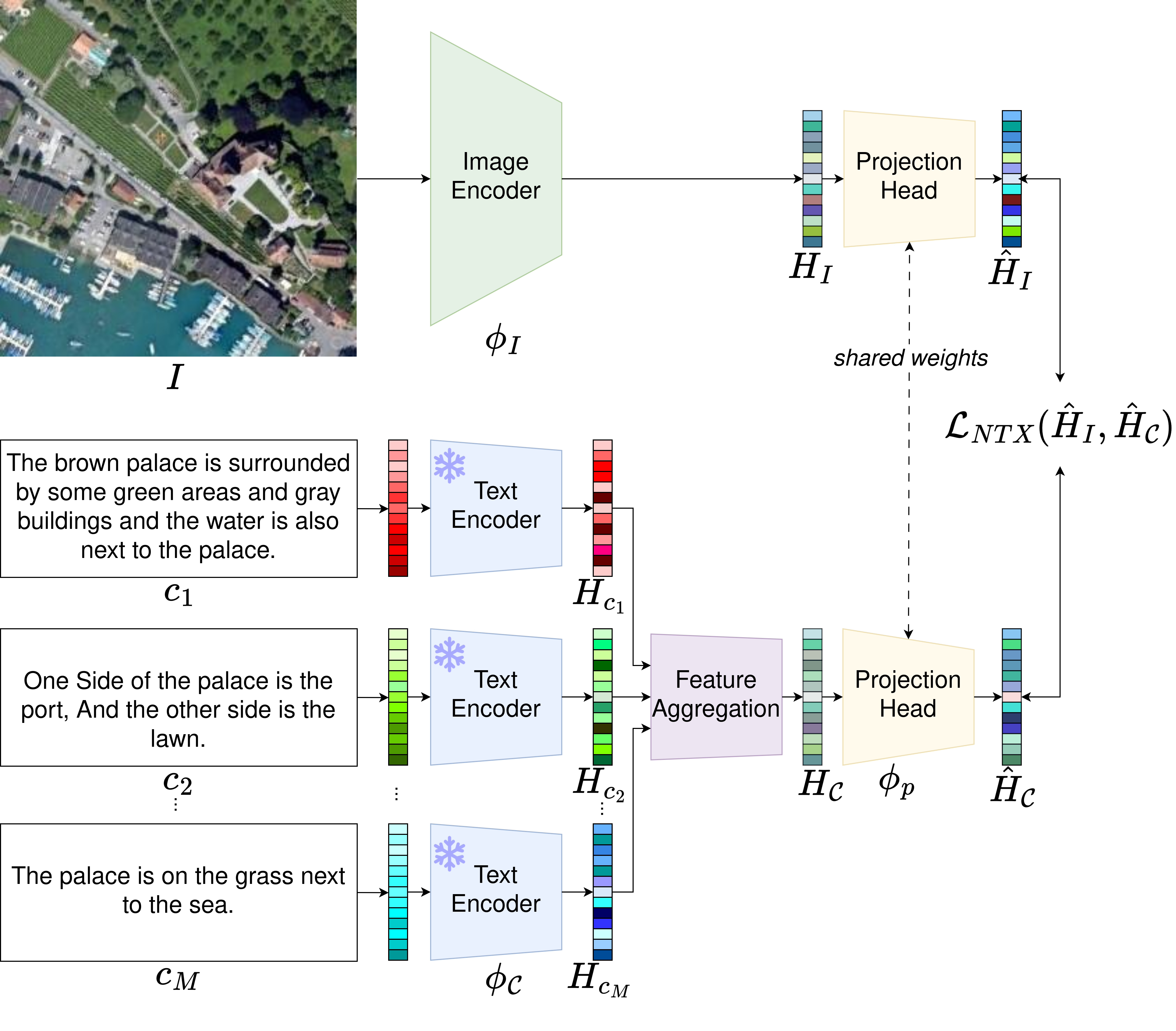}
    \captionsetup{belowskip=\bskip}
    \caption{
    An illustration of the proposed \acf{wfa} strategy for pretraining the considered \ac{vlm} (which consists of an image encoder $\phi_I$, a text encoder $\phi_\mathcal{C}$, and a projection head $\phi_p$ shared for image and text) with \ac{ntxent} loss $\mathcal{L}_{\text{NTX}}$. The proposed WFA strategy exploits complementary information from all captions of an image at once, while reducing redundancies through feature aggregation with importance weighting.
    }
    \label{fig:multimodel}
\end{figure}

To calculate adaptive importance weights of captions in $\mathcal{C}$ (e.g., $\alpha_j$ for $c_j$), we propose two techniques: i) non-parametric uniqueness; and ii) learning-based attention.
The weighting factor $\alpha_j$ is estimated on the basis of the proposed techniques. By increasing or decreasing the contributions of individual captions, the proposed techniques aim to maximize information content and effectively reduce redundancies. The proposed techniques are explained in detail in the following, and their overview is shown in \Cref{fig:multimodel}.

\subsubsection{Non-Parametric Uniqueness} 

This technique aims to measure the importance of captions in $\mathcal{C}$ based on their uniqueness compared to each other. 
We assume that uniqueness of captions can be estimated based on their linguistic distinction to each other. 
To this end, we utilize the widely-used \ac{bleu} \cite{10.3115/1073083.1073135} score for measuring linguistic alignment, i.e., the opposite of distinction. 
In detail, we define a uniqueness score $\uni(\mathcal{C},j)$ for caption $c_j$ based on the \ac{bleu} score with respect to all other captions of $\mathcal{C}$. 
Specifically, we use \ac{bleu}-4 for this calculation.
The uniqueness of the text $c_j$ is its \ac{bleu}-4 score subtracted from one as follows:
\begin{align}
    \uni(\mathcal{C},j)        &= 1-\bleu_4(c_j, \mathcal{C} \setminus c_j).
\end{align}
where $\bleu_4(c_j, \mathcal{C} \setminus c_j)$ is the \ac{bleu}-4 score of the $j$th caption in the set of texts $\mathcal{C}$ against all other texts $\mathcal{C} \setminus c_j$.
The result is normalized using the softmax operation to obtain the weighting factors $\alpha_j^{\uni}$ as:

\begin{equation}
    \begin{aligned}
        \alpha_j^{\uni}  &= \sigma(\boldsymbol{\uni})_j \\
                     &= \frac{\exp{\uni(\mathcal{C},j)}}{\sum_{k=1}^{M}\exp{\uni(\mathcal{C},k)}}.
    \end{aligned}
\end{equation}
 where $\boldsymbol{\uni} = \langle\uni(\mathcal{C},1), \uni(\mathcal{C},2), \dots, \uni(\mathcal{C},M)\rangle$ is the vector of all uniqueness scores.
The weighting factors can be used as in (\ref{eq:weighted_avg}) to get the aggregated features as follows:
\begin{equation}
    \begin{aligned}
        H_\mathcal{C}^{\uni} &= \sum_{j=1}^M \alpha_j^{\uni}\cdot H_{c_j}\\
               &= \sum_{j=1}^M \sigma(\uni(\mathcal{C},j))_j\cdot H_{c_j}.
    \end{aligned}
\end{equation}
Since $\uni(\mathcal{C},j)$ is non-parametric and unlearned, it is computationally efficient to estimate (i.e., it can be calculated offline and re-used in different iterations of the pretraining process).

\subsubsection{Learning-based Attention}
This technique aims to estimate a weighting factor $\alpha_j^{\att}$ for caption $c_j$ based on the caption features $H_{c_j}$.
To this end, $H_{c_j}$ is transformed via multiple transformer encoder blocks \cite{NIPS2017_3f5ee243} and an additional linear layer into a pseudo-weight $\alpha_j^\prime$ as follows:
\begin{equation}
    \begin{aligned}
        \alpha_j^\prime &= \att(H_{c_j}) \\
                    &=\FC\left(\TE\left(H_{c_j}\right)\right),
    \end{aligned}
\end{equation}
where $\TE$ is a transformer encoder with two layers as defined in \cite{NIPS2017_3f5ee243} and $\FC$ is a fully-connected layer followed by a non-linearity.
The pseudo-weights $a_j^\prime$ are calculated for all text embeddings $H_{c_j},{j = 1,2,\dots,M}$ independently.
We would like to note that this technique uses intra-caption attention only to calculate weights.
This means that it uses attention on every caption individually. 
Therefore, this technique is independent of the number of captions within the set of texts for image $I$ and allows for different images to have different numbers of associated captions.
The final weighting factors $\alpha_j^{\att}$ are then calculated by applying the softmax operation over all pseudo-weights as follows:
\begin{align}
    \alpha_j^{\att} &= \sigma(\boldsymbol{\alpha^\prime})_j,
\end{align}
where $\boldsymbol{\alpha^\prime} = \langle\alpha_1^\prime, \alpha_2^\prime, \dots, \alpha_M^\prime\rangle$ is the vector of all pseudo-weights. 
Therefore, the aggregated text feature $H_{\mathcal{C}}^{\att}$ can be obtained as follows:
\begin{equation}
    \begin{aligned}
        H_{\mathcal{C}}^{\att} &= \sum_{j=1}^M \alpha_j^{\att} \cdot H_{c_j}\\
              &= \sum_{j=1}^M \sigma(\att(H_{c_k})|_{k = 1,2,\dots, M})_j \cdot H_{c_j}.\\
    \end{aligned}
\end{equation}

After obtaining the aggregated text features, they are fed into the projection layer together with the image features as follows: $\hat{H}_{\mathcal{C}} = \phi_p(H_{\mathcal{C}})$ and $\hat{H}_I = \phi_p(H_I)$. 
Then, for pretraining the considered \ac{vlm} model, the resulting representations are used for image-text alignment through $\mathcal{L}_{\text{NTX}}(\hat{H}_{\mathcal{C}}, \hat{H}_I)$ ~\cite{sohn2016improved}. 
The pretrained \ac{vlm} model can be used to obtain $\hat{H}_I$ and, depending on the selected technique, $\hat{H}_\mathcal{C}^{\uni}$ or $\hat{H}_\mathcal{C}^{\att}$, which can be used for downstream tasks.

\section{Experimental Results}
\new{In the experiments, for the VLM pretraining, we combine the following four image-text datasets: i) GAIA  \cite{zavras2025gaiaglobalmultimodalmultiscale}; ii) RSICD \cite{8240966}; iii) RSITMD \cite{9437331}; and iv) NWPU-Captions \cite{cheng2022nwpu}.
The resulting pretraining dataset comprises 88,194 images with 440,970 captions.
We divide this dataset into training (90\%) and validation (10\%) subsets.}

\begin{table*}
	\centering
	\caption{Text-to-image retrieval results on UCM and \new{Sydney} in terms of BLEU-4 (\%), \ac{map} (\%) and \acs{flops} (B) obtained by different \ac{vlm} strategies. The results are obtained for top-5 and top-20 retrieved images. \Ac{lfa} indicates whether the considered importance weighting technique introduces additional learned parameters or not. \ours{}~indicates our strategy and techniques. \new{Best results for \ac{lfa}/no-\ac{lfa} are given in bold.}}
\setlength{\tabcolsep}{8pt}
\begin{tabular}{@{}lcccccccccc@{}}
	\toprule
	
	\multirow{2}{*}{\thead{\textbf{Strategy}}} &
	\multirow{2}{*}{\thead{\textbf{LFA}}} &
	\multicolumn{2}{c}{\textbf{BLEU-4}@5(\%) $\uparrow$} &
    \multicolumn{2}{c}{\textbf{mAP}@5(\%) $\uparrow$} &
	\multicolumn{2}{c}{\textbf{BLEU-4}@20(\%) $\uparrow$} &
    \multicolumn{2}{c}{\textbf{mAP}@20(\%) $\uparrow$} &
	\multirow{2}{*}{\textbf{FLOPS (B) $\downarrow$}} \\
	
	& & \multicolumn{1}{c}{UCM} & \multicolumn{1}{c}{Sydney} & \multicolumn{1}{c}{UCM} & \multicolumn{1}{c}{Sydney} &  \multicolumn{1}{c}{UCM} & \multicolumn{1}{c}{Sydney} & \multicolumn{1}{c}{UCM} & \multicolumn{1}{c}{Sydney} &  \\
	\midrule
	
	Replication & \multirow{5}{*}{\xmark} & 27.8 & 25.9 & 49.6 & 22.9 & 28.2 & 26.5 & 48.6 & 23.0 & 171.92\\
	
	Concatenation & & 13.6 & 11.3 & 30.1 & 31.3 & 16.2 & 8.0 & 33.0 & 30.0 & 34.40\\
	
	Random Selection & & 23.8 & 30.5 & 44.6 & 58.7 & 24.8 & 28.7 & 44.6 & 58.2 & \textbf{34.38}\\
	
	Mean Feature \cite{multiqueryvideo} & & 35.6 & 15.4 & 62.5 & 34.9 & 33.2 & 11.3 & 62.4 & 33.4 & 35.80\\
	\ours{WFA} (Uniqueness) & & \oursnumber{\textbf{40.8}} & \oursnumber{\textbf{40.6}} & \oursnumber{\textbf{67.5}} & \oursnumber{\textbf{88.0}}  & \oursnumber{\textbf{42.3}} & \oursnumber{\textbf{41.0}} & \oursnumber{\textbf{69.3}} & \oursnumber{\textbf{85.6}} & \oursnumber{35.80}\\\midrule
	
	LG-WF \cite{multiqueryvideo} & \multirow{2}{*}{\cmark} & 38.4 & 21.0 & 64.1 & 42.0 & 37.8 & 21.7 & 65.6 & 42.1 & \textbf{35.80} \\
	
	\ours{WFA} (Attention) & & \oursnumber{\textbf{49.9}} & \oursnumber{\textbf{44.2}} & \oursnumber{\textbf{82.0}} & \oursnumber{\textbf{87.6}} & \oursnumber{\textbf{45.8}} & \oursnumber{\textbf{42.8}} & \oursnumber{\textbf{78.2}} & \oursnumber{\textbf{86.3}} & \oursnumber{36.47}\\
	\bottomrule
\end{tabular}

\label{tab:all_results}
\end{table*}

As the image encoder of the considered \ac{vlm}, we utilize the \new{DOFA Base model \cite{xiong2024neuralplasticityinspiredmultimodalfoundation} pretrained on RS data} with the input image size of $224 \times 224$. As the text encoder, we utilize a small variant of the pretrained \ac{bert} \cite{turc2019well} model.
For text-level embeddings from \ac{bert}'s token-level embedding space, we average the sum of the last four hidden states following \cite{mikriukov2022deep}.
We utilize a \ac{mlp}-based projection head with one hidden layer. The text encoder weights are kept frozen during pretraining, while we fine-tune the image encoder, projection head and, for the learned technique, weights for weighted averaging.
We pretrain the considered \ac{vlm} for a maximum of 1000 epochs by using the \acs{adamw} \cite{loshchilov2018decoupled} optimizer and early stopping with a patience of 5 epochs based on the BLEU-4 score on the validation set. 
The learning rate, weight decay, temperature and batch size are set to \new{10$^{-5}$, 10$^{-4}$, 0.5, and 200,} respectively. VLM pretraining of the considered model takes less than 24 hours on a single NVIDIA H100 GPU.

We compare the results of our \ac{wfa} strategy with those obtained by five different strategies to exploit various captions in VLM training: i) Replication (which replicates images to match the number of captions); ii) Concatenation (which concatenates multiple captions of an image as a single text); iii) Random Selection (which randomly selects one caption per image), iv) Mean Feature (which averages the text features) \cite{multiqueryvideo}; and v) \ac{lgwf} (which learns the weights to merge text features through a single linear layer) \cite{multiqueryvideo}.
To the best of our knowledge, all these strategies, except Replication, are applied for the first time in the context of \acp{vlm} in \ac{rs}. 

In the experiments, as a downstream task we consider text-to-image retrieval. 
To this end, we exploit the UCM-captions \cite{qu2016deep} and Sydney-captions \cite{qu2016deep} datasets. 
Each image within the considered datasets is associated with five captions, resulting in 10,500 captions for UCM-captions and 157,500 captions for Sydney-captions. 
Each caption of each image is selected as a query text, while image retrieval is applied to the respective dataset. 
Results of each strategy are provided in terms of: i) \ac{flops}, ii) \ac{bleu}-4; and iii) \ac{map} scores. 
The scores are obtained in 10,500 trials performed with 10,500 selected query texts from the UCM-captions dataset, while they are attained in 157,500 trials performed with 157,500 selected query texts from the Sydney-captions dataset. 
The retrieval performance was assessed on the top-5 and top-20 retrieved images.
It is worth noting that the datasets considered in the evaluation were neither used during pretraining nor for fine-tuning. 
Accordingly, the retrieval performance can be considered as zero-shot performance.
We take into account only the inference complexity to calculate the \ac{flops} for each strategy. 
It is worth emphasizing that \ac{flops} are input-size dependent and the considered datasets have varying caption lengths.
Therefore, we consider the average caption length for \ac{flops} calculation in our computational complexity assessment.

\Cref{tab:all_results} shows the results of the proposed \ac{wfa} with non-parametric uniqueness technique (denoted as \ac{wfa} (Uniqueness)), the proposed \ac{wfa} with learning-based attention technique (denoted as \ac{wfa} (Attention)) and all the other strategies used for the comparison for the UCM and \new{Sydney} datasets. 
In the table, we group the strategies into \ac{lfa} strategies (which introduce additional learned parameters to the \ac{vlm}) and no-\ac{lfa} strategies (which do not introduce additional learned parameters to the \ac{vlm}).
From the table one can observe that
\ac{wfa} (Attention) in general achieves the highest \ac{bleu}-4 and \ac{map} scores at the cost of a slight increase in FLOPS for both datasets. 
For  example, it outperforms the Replication strategy by 32.4\% and 65.3\% in \ac{map}@5 for the UCM and Sydney datasets, respectively. 
In the case of \ac{map}@20, the improvement with respect to the Replication strategy is 29.6\% and 53.3\% for the UCM and Sydney datasets, respectively. 
Among no-\ac{lfa} approaches, \ac{wfa} (Uniqueness) demonstrates the best performance. 
As an example, it yields improvements of 20.7\% and 52.6\% in \ac{map}@20 compared to Replication for the UCM and Sydney datasets, respectively. 
In addition, \ac{wfa} (Uniqueness) also surpasses the \ac{lfa}-based \ac{lgwf} strategy for both datasets.
Concerning computational complexity, the Replication strategy has the highest complexity with \new{171.92} billion \ac{flops}. 
This is due to its requirement to apply a forward pass for every sample, which includes the computations for both the image and text encoders.
Therefore, to process all five captions that correspond to one image, the image and text encoders have to be used five times each.
In contrast, all other strategies including our \ac{wfa} need only a single forward pass per image input, as the extracted image features can be reused between text inputs.
\new{They all require \ac{flops} only in the range of 34.38 -- 36.47 billion.}

\section{Conclusion}

In this paper, we have presented \ac{wfa}, a strategy for \ac{vlm} pretraining in \ac{rs} that uses feature aggregation of captions based on importance estimation.
In addition, we have introduced two importance estimation techniques: i) non-parametric uniqueness; and ii) learning-based attention that calculate adaptive weights for different captions.
Our techniques optimize the text inputs during pretraining by enhancing their unique features, while reducing redundant content. 
Thus, they enable more efficient and effective \ac{vlm} pretraining in \ac{rs}.
We have compared the proposed \ac{wfa} strategy and techniques with five pretraining strategies, focusing on the text-to-image retrieval task as a downstream task.
The experimental results demonstrate the success of our strategy and techniques. 
In particular, we would like to highlight that, compared to the Replication strategy (which is widely used for \ac{vlm} pretraining in \ac{rs}), our strategy significantly reduces the pretraining and inference time, while improving the text-to-image retrieval accuracy.  
Based on our experimental analysis, we have derived some guidelines as follows:
\begin{enumerate}
    \item If sufficient computational resources are available, the \ac{wfa} strategy with the learning-based attention technique is suggested to be selected, as it is the best-performing approach among all the considered ones.
    \item If the compute resources are limited and the pretraining set is highly redundant, the \ac{wfa} strategy with the Uniqueness technique can be selected. This strategy has lower computational requirements (since it relies on the \ac{wfa}-weights computed before the training process) and results in a good downstream performance.
    \item The widely used Replication strategy shows inferior performance compared to other strategies, while having a high computational cost. Thus, we suggest not to use it.
\end{enumerate}

As future developments of our work, we plan to: i)~evaluate the performance of our techniques under different downstream tasks, including visual question answering and captioning; and ii)~scale up \ac{vlm} pretraining with more data and deeper architectures. 

\FloatBarrier
\small
\printbibliography

\end{document}